\documentclass{article}

\usepackage{arxiv}
\usepackage[utf8]{inputenc}
\usepackage[T1]{fontenc}
\usepackage{amsfonts}
\usepackage{hyperref}
\usepackage{indentfirst}
\usepackage{cite}
\usepackage{graphicx}
\usepackage{subfigure}

\title{CropDefender: deep watermark which is more convenient to train and more robust against cropping}

\author{Jiayu Ding\\
Xi'an Jiaotong University\\
\texttt{jiayu.ding@foxmail.com}
\And
Yuchen Cao\\
Zhejiang University\\
\texttt{3180101476@zju.edu.cn}
\And
Changhao Shi\\
Xi'an Jiaotong University\\
\texttt{2535450187@qq.com}
}

\begin{document}

\maketitle
\begin{abstract}
Digital image watermarking, which is a technique for invisibly embedding information into an image, is used in fields such as property rights protection. In recent years, some research has proposed the use of neural networks to add watermarks to natural images. We take StegaStamp\cite{Tancik_2020_CVPR} as an example for our research. Whether facing traditional image editing methods, such as brightness, contrast, saturation adjustment, or style change like 1-bit conversion, GAN, StegaStamp has robustness far beyond traditional watermarking techniques, but it still has two drawbacks: it is vulnerable to cropping and is hard to train. We found that the causes of vulnerability to cropping is not the loss of information on the edge, but the movement of watermark position. By explicitly introducing the perturbation of cropping into the training, the cropping resistance is significantly improved. For the problem of difficult training, we introduce instance normalization to solve the vanishing gradient, set losses' weights as learnable parameters to reduce the number of hyperparameters, and use sigmoid to restrict pixel values of the generated image.
\end{abstract}

\section{Introduction}
Digital image watermarking is a technique for embedding information into an image without disturbing the visual appearance of the image. The most common application of digital image watermarking is property rights protection, where the creator can hide authorship information in the image and decode the information when proof is needed. In addition, the technique can also be used as an encryption method, as long as the encoding and decoding methods are covert. To be able to achieve these purposes, imperceptibility and resistance to perturbation are the two main requirements of watermarking technology.

Traditional image watermarking techniques include the least significant bits (LSB) and frequency domain watermarking. LSB hides information by displacing the least significant bits of the original image pixels, which has no effect on the visual appearance of the image, but the slightest perturbation with the image can corrupt the information. Frequency domain watermarking, although more robust than LSB, can also be corrupted by Gaussian blurring, etc.

In recent years, some research has started to use neural networks to add watermarks. For example, StegaStamp\cite{Tancik_2020_CVPR}, which has the best robustness to date, uses an encoder network to transform the original image with the information into a watermarked image, and then uses a decoder network to retrieve the information, with differentiable perturbation inserted between the encoder and decoder during training to increase the robustness. In our experiments, we found that the neural-network-based watermark generation technique has amazing robustness. Regardless of whether the encoded image undergoes traditional image editing methods, such as brightness, contrast, saturation adjustment, or style change like 1-bit conversion, GAN, the decoding has a very high accuracy rate. 

Although StegaStamp has demonstrated strong robustness in real-world scenarios, it still suffers from two drawbacks: it is not easy to train and it is not resistant to cropping. The lack of ease of training means that users have difficulty in obtaining their own encoders and decoders and have to use pre-trained models, which reduces the security of encryption. And since cropping is a common image processing in daily life, the lack of cropping resistance reduces the generalizability of encryption.

For the crop resistance, it is experimentally demonstrated that the main factor leading to the accuracy reduction is not the loss of information, but the movement of the watermark position, which means that the decoder's convolutional neural network implicitly captures the position information. By explicitly introducing the perturbation of cropping into the training, the cropping resistance is significantly improved.

The key reason for the difficulty of training is that image fidelity and information fidelity are contradictory, and information fidelity is much more difficult to learn than image fidelity, which makes the model easy to learn only image fidelity and ignore information fidelity, so two types of losses must be well balanced.To solve this problem, StegaStamp uses a large number of hyperparameters to change losses' weights during training. By taking losses' weights as learnable parameters, we can achieve more automatic and smoother model training by setting only initial weights. StegaStamp also has a high dependence on model parameter initialization. In most cases, the model quickly falls into a suboptimal region that is difficult to escape from, and the decoding accuracy stays at a very low level. By adding instance normalization, our model can reach a high accuracy rate steadily in the initial stage. Furthermore, in order to keep the encoder from avoiding perturbation by generating watermarks far beyond the 0-1 range, we use sigmoid to restrict the generated pixels to conform to the value domain, which improves the stability of the training, although it affects the image quality to some extent.

Our model explores for the first time how image watermark generators can be trained more consistently and conveniently.  We also experimentally explore in depth the amazing robustness of deep watermarking and improve the crop resistance of deep watermarking, making it possible to be applied to a wider range of scenarios.

\section{Related Work}

\subsection{Digital Image Watermarking}
Digital watermarking technology refers to embedding hidden marks in digital multimedia data by signal processing method. This mark is usually invisible and can be extracted only through a special detector or reader. Digital watermarking is an important research direction of information hiding technology.For typical scenarios, the watermarking process can be roughly divided into three stages: embedding, attack and decoding\cite{info11020110}. 

Among previous works lies LSB steganography (least significant bit steganography) that hides changes that is imperceptible to the human eye.Also in frequency domain such as DCT, DFT, DWT, SVD etc, the watermark is inserted into transformed coefficients of image giving more information hiding capacity and more robust \cite{2013A,2020A,SARKAR2015DIGITAL}

\subsection{Deep Watermarking}
There has been some research working on using neural network to hiding information into the image. Qiang Wei et al. use cycle variational autoencoder to embed QR codes into human face images\cite{Wei2020ARI}. Shumeet Baluja works on hiding color images into any natural images\cite{NIPS2017_838e8afb}. These methods embed lots of information into the image, but they don't focus on watermark's robustness against perturbations. By using error correcting code and explicitly introducing differentiable pertubations between encoding and decoding during training, StegaStamp\cite{Tancik_2020_CVPR} can robustly embed bit string into the image, although the number of embedded bits is much less than methods before. 

\section{Implementation Details}
The basic structure of the network is encoder-transformation-decoder. First, the encoder embeds a binary string into an image to form a watermarked image. Then, the transformation part simulates real world situation by random perturbations. Finally, the decoder retrieves the binary string from the watermarked image. 

Looking into the details, in StegaStamp, activation layers directly follow convolutional layers or fully connected layers, which brings difficulty to deep neural network training. By inserting instance normalization layers \cite{DBLP:journals/corr/UlyanovVL16} before activation layers, we make the network easier to train. 

For training, images are from the MIRFLICKR25k dataset\cite{10.1145/1460096.1460104}, and binary strings are randomly generated. The optimizer used is Adam\cite{DBLP:journals/corr/KingmaB14}. For testing, we sample 1020 images from caltech101 dataset\cite{FeiFei2004LearningGV} as test dataset.
\subsection{Encoder}
Similar to StegaStamp, the input of the encoder is a $400\times400\times3$ image $I_{input}$ and a 100 bit binary string (message). The binary string is first turned into a $50\times50\times3$ tensor by a fully connected layer, then upsampled to a $400\times400\times3$ tensor. The message tensor and the image tensor are merged to sent into a U-Net\cite{DBLP:journals/corr/RonnebergerFB15} style encoder. The output of the encoder is a $400\times400\times3$ residual tensor $R$. To get the watermarked image $I_{encoded}$, StegaStamp just adds the residual and the input image together:
$$I_{encoded}=I_{input}+R$$
but our model uses sigmoid function $\sigma$ to force the pixel value of the watermarked image in range [0,1] (-0.5 for zero mean):
$$I_{encoded}=\sigma(I_{input}-0.5+R) $$
\subsection{Transformation}
Between the encoder and the decoder, a set of differentiable image perturbations are applied to the watermarked image to simulate real world situation, including perspective warp, motion and defocus blur, color manipulation and noise. The implementation details could be seen in the paper of StegaStamp\cite{Tancik_2020_CVPR}. Our improvement is introducing cropping into the training. The area and the aspect ratio of the cropped image are both random, and the cropped image will be resized to $400\times400$ to be sent to decoder. A trick for convergence is to gradually increase the strength of perturbations, so that the training will not be too difficult at the beginning .
\subsection{Decoder}
For decoding, the transformed image first goes through a spatial transformer network \cite{DBLP:journals/corr/JaderbergSZK15} for robustness against perspective warp. Then it will be sent to a convolutional network (with fully connected layer at the end) for decryption. The output of the decoder is a 100 bit binary string, which is expected to be the same as the input message.
\subsection{Loss}
There are two requirements for our model: image and information fidelity. For image fidelity, $L_2$ residual regularization $L_R$ and LPIPS perceptual loss \cite{Zhang_2018_CVPR} $L_P$ are used to describe the similarity between the watermarked image and the original image. For information fidelity, we use classic cross entropy loss $L_M$. In StegaStamp, the training loss is the linear combination of these losses:
$$L=\lambda_R L_R+\lambda_P L_P+\lambda_M L_M$$
The weights $\lambda_R$ and $\lambda_P$ are set to zero before 1500 steps, and then linearly increase to the max value to aid in convergence. This introduces too many hyperparameters into training. In our model, we use self adaptive weights \cite{Kendall_2018_CVPR} to reduce the number of hyperharameters:
$$L=\frac{L_R}{\sigma_R^2}+\frac{L_P}{\sigma_P^2}+\frac{L_M}{\sigma_M^2}+2log(\sigma_R\sigma_P\sigma_M)$$
$\sigma_{R,P,M}$ here are all learnable parameters. Only the initial values of $\sigma$s need to be determined. If one $\sigma$ increases, the corresponding loss term will decrease, but the corresponding log term will increase, and vice versa.It forms an automatic balance between different losses, and sure makes the training process smoother and better.

\section{Robustness}

To explore the robustness and weaknesses of deep watermark, we conducted a series of experiments where we embedded 100-bit messages for test dataset, performed various edits on the encoded images, and then decoded them. We then found deep watermark surprisingly robust in most of the experiments, whereas it has some unexpected weakness as well.
\subsection{Image Enhancement}
\begin{figure}
\centering 
\subfigure{
\label{EnhanceFigure1}
\includegraphics[width=0.45\textwidth]{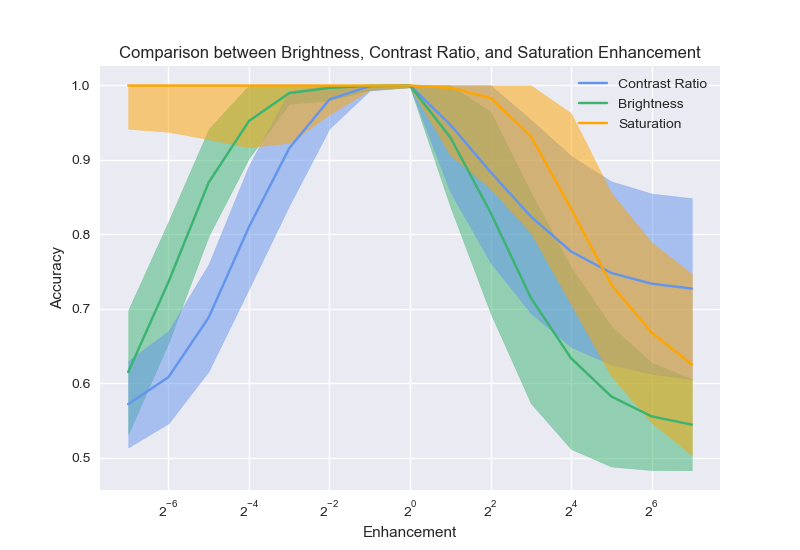}}
\subfigure{
\label{EnhanceFigure2}
\includegraphics[width=0.45\textwidth]{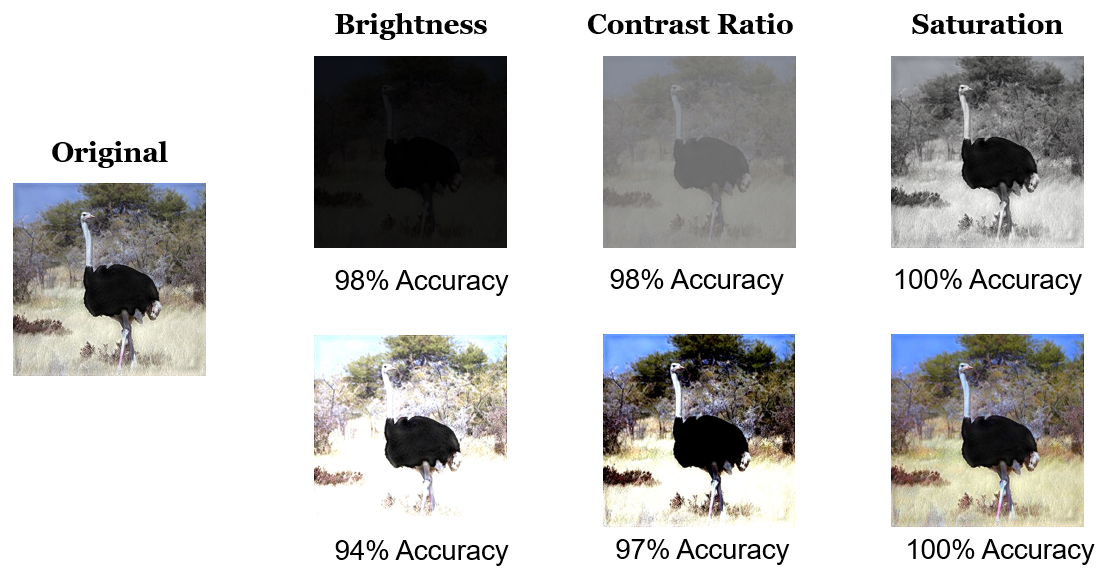}}
\caption{The result of modifying brightness, contrast ratio, and saturation on a data set of 1000 images}
\label{EnhanceFigure}
\end{figure}
The adjustment of brightness, contrast ratio and saturation is common in daily life. Figure \ref{EnhanceFigure1} shows the effect of these traditional image editing methods on the decode accuracy. When the changes are not too drastic, the average accuracy is over 90 percent, which is quite impressive. For example in Figure \ref{EnhanceFigure2}, an image with low brightness, which is close to pure black, also has accuracy of 98 percent, and an image with low saturation, which is close to gray scale, still has accuracy of 100 percent.

An interesting thing is that decreasing the saturation does not affect the decoding accuracy. In fact, there is no decrease in accuracy even if the image is converted to gray scale. Similarly, when increasing the contrast ratio, there is a lower bound of decoding accuracy at around 70 percent. This is because the decrease of the saturation and the increase of the contrast ratio cannot turn an image into a pure color image, while other adjustments can.

\subsection{Style Change}
\begin{figure}
    \centering
    \subfigure{
    \label{violinFigure1}
    \includegraphics[width=0.45\textwidth]{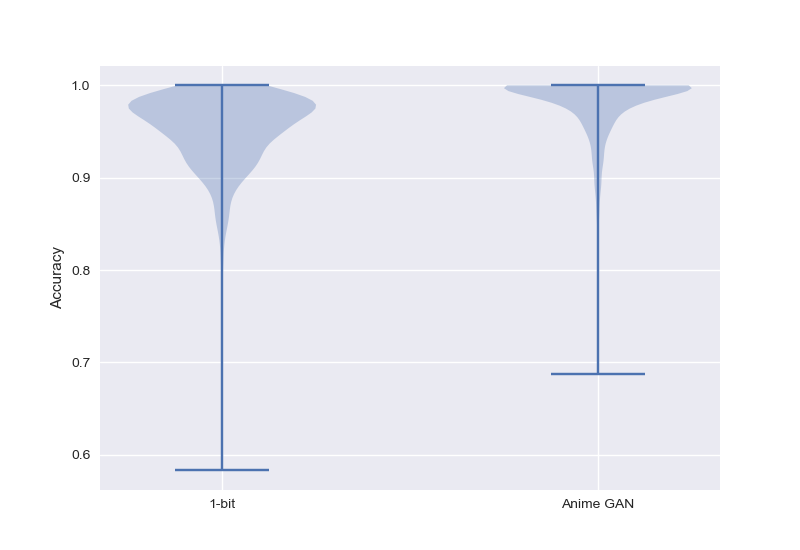}}
    \subfigure{
    \label{violinFigure2}
    \includegraphics[width=0.45\textwidth]{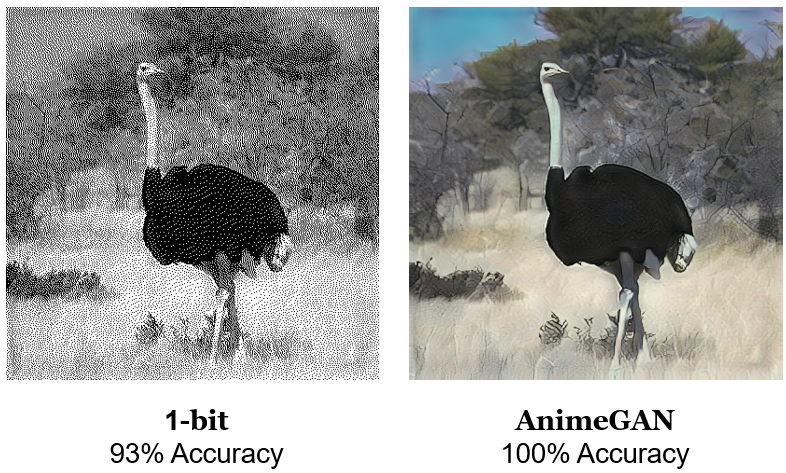}}
    \caption{The result of  1-bit, and  GAN on a data set of 1020 images.Histogram equalization and Edge enhance have almost no effect on accuracy of stagastamp, while it also performs well with 1-bit conversion and GAN modification.}
    \label{violinFigure}
\end{figure}
Different with image enhancement which remains most of the information of the image, we have tried some methods that make significant changes to the style of the images, such as GAN and 1-bit conversion(Figure \ref{violinFigure}).

For 1-bit conversion,the convert image contains only black and white pixels, and the different density of black pixels makes it look like a grayscale image. We find that even after such a loss of information, the result still remains an average accuracy of 95 percent.

For GAN, we test AnimeGAN, a kind of GAN that can convert pictures into animation style. It can be seen clearly that perceptual features have been strongly influenced by AnimeGAN,whereas the accuracy of most pictures still stays at 100 percent, which means almost no information of the watermark has been influenced in the process of style-transfering.

\section{Model Enhancement}
\subsection{Cropping Resistance}

\begin{figure}
    \centering
    \includegraphics[width=1\textwidth]{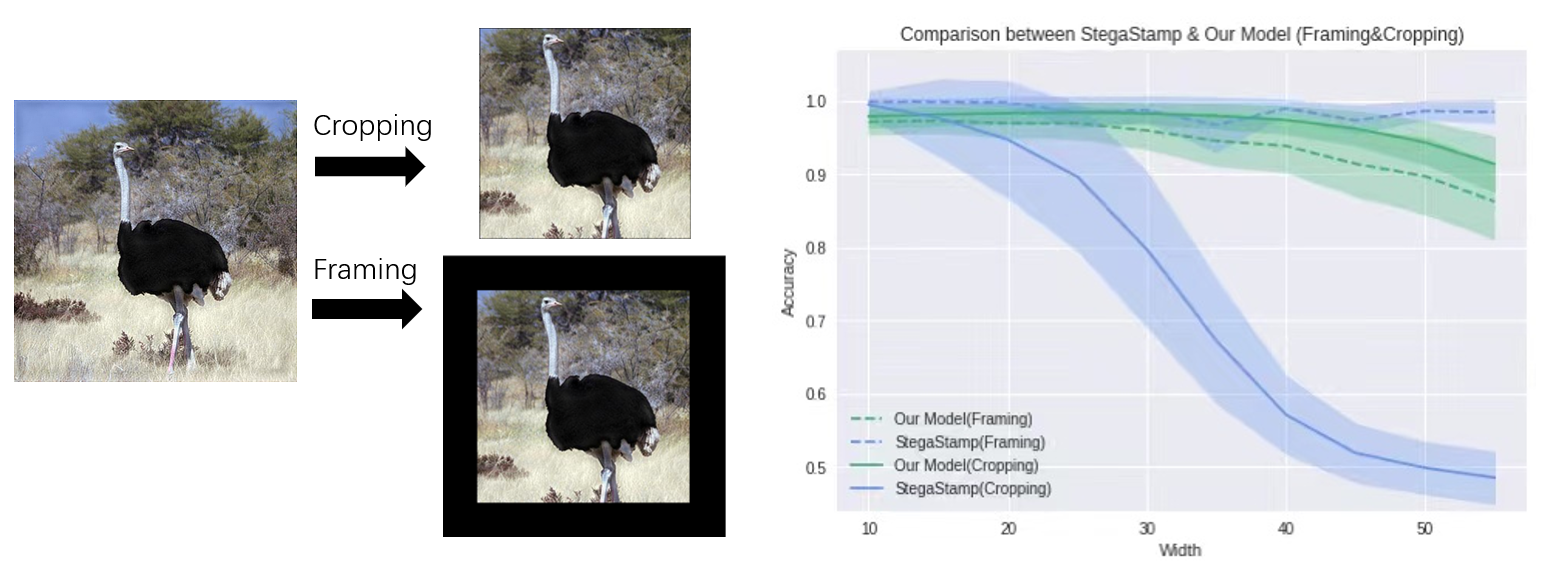}
    
    \caption{The comparison on the effect of center-cropping and framing on accuracy between StagaStamp and Our Model. The pictures are processed in the same way with StegaStamp and Our Model, and are processed with Cropping and Framing respectively.}
    \label{croppingframing}
\end{figure}

\subsubsection{Experiments on StegaStamp}
Among all the experiments we conducted on StegaStamp, there was one specific weakness that we detected, that StegaStamp is vulnerable with cropping.When each picture is cropped, each edge must be cut at the corresponding width.(Figure \ref{croppingframing}) While StegaStamp shows its robustness with other approaches of processing, cropping destroys the watermark on a large degree. It can be seen that when the width of cropping increase to 30 pixels, the decoder can only restore 80 percent of the original message. At around the width of 50 pixels, the accuracy hit 50 percent, which means that the message cannot be decoded from the image at all.

To figure out whether the impact was born from the loss of information in the process of cropping or not, we conducted a relatively similar experiment with framing. We add black borders of different widths to the image and then decode it. It can be seen that the accuracy stays above 95 percent, even when the width is increased to 70, which means 57.75 percent of the pixels are covered, which differed drastically from the results of cropping. The results prove the fact that the decrease in accuracy didn't come from the loss of information in the process of cropping, which shed some light on the method that we adopted to improve the model.

As is demonstrated by our experiment, the weakness of cropping derives from the movement of the watermark position, which means that the decoder’s convolutional neural-network implicitly captures the position information.
\subsubsection{Our Improvement}
In order to ameliorate the cropping resistance of the model, we explicitly introduced the perturbation of cropping into the training. And with the retrained model, we conducted the same experiment to figure out the progress. With our model, the accuracy stays above 90 percent when the crop width is 30 pixels, which means 27.7 percent of the picture is cropped. Even at the width of 50 pixels, when the accuracy of Stegastamp decreases to 50 percent, our model still remains 88 percent of the accuracy. As for framing, compared with the original model, the accuracy of our model for the frame is slightly reduced, but it still remains at a high level of 90 percent, and the degree of accuracy decline is within an acceptable range. In that perspective, by include the perturbation that hasn't been introduced into StegaStamp, we successfully solved the problem that the original model is vulnerable with cropping.

\subsection{Ease of Training}
\subsubsection{Instance Normalization}
\begin{figure}
    \includegraphics[width=\textwidth]{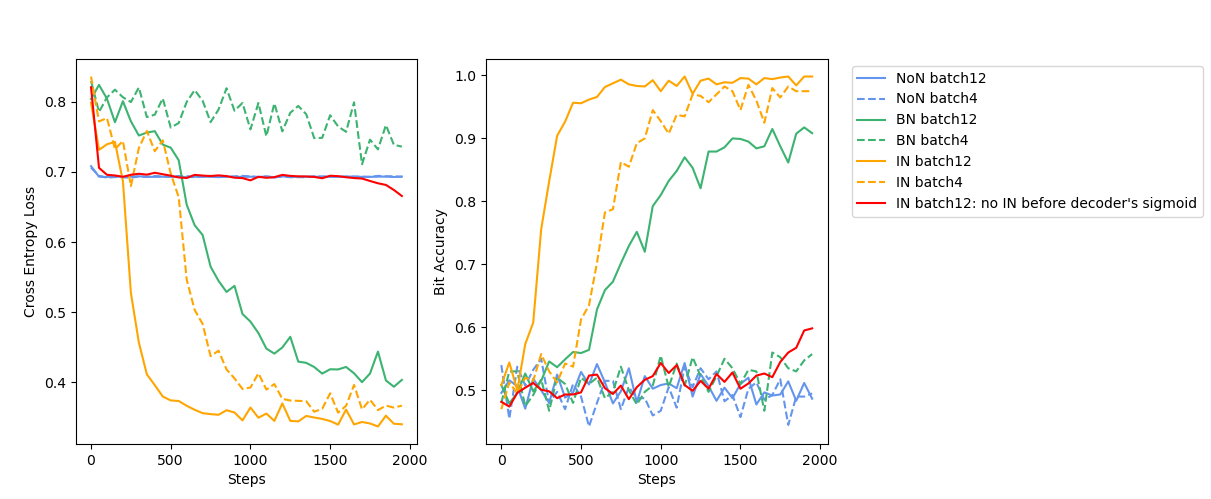}
    \caption{The impact of normalization on accuracy and cross entropy loss. "NoN" means no normalization layer in the network; "BN" means the batch normalization layer is inserted before every activation layer; "IN" means the instance normalization layer is inserted before every activation layer; "decoder's sigmoid" means the last activation layer of the decoder, which uses sigmoid. The cross entropy loss and bit accuracy is calculated on each batch.}
    \label{lossFigure}
\end{figure}
In our experiments, we found that StegaStamp easily falls into "bad regions" at the beginning of training, which means that decoding accuracy is always around 50\%,and the cross entropy loss is high (Figure \ref{lossFigure}).  Even after 100k steps, the model cannot escape these "bad regions".StegaStamp can only rely on luck to get good parameters at initialization to keep it from falling into "bad regions". It is important to achieve high decoding accuracy early in the training process, because the strength of perturbations and the difficulty of training gradually increase.  If the model cannot achieve high decoding accuracy early on, it will be more difficult to learn how to decode later. Moreover, it has been shown that the early stage of training has a critical impact on the final result. \cite{DBLP:journals/corr/abs-1711-08856}
\begin{figure}
    \centering
    \includegraphics[width=.546\textwidth]{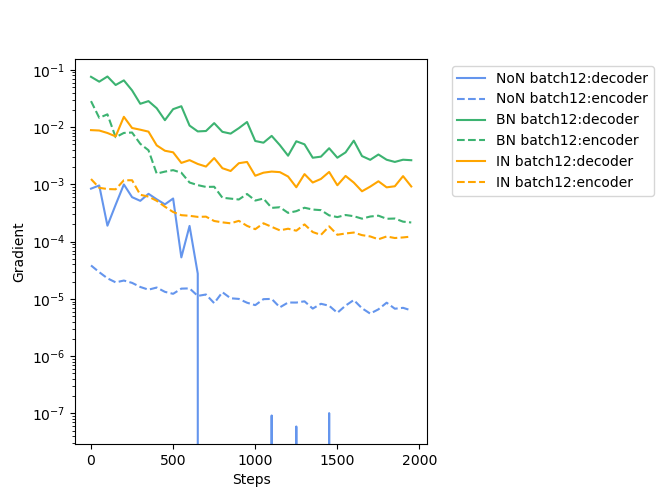}
    \caption{The impact of normalization on gradients. We use the average of the absolute values of the gradients to represent the gradients of the whole layer. "encoder" means the fully connected layer changing the binary string to $50\times50\times3$ tensor; "decoder" means the first convolutional layer in the decoder's decryption part. }
    \label{gradFigure}
\end{figure}

The difficulty of escaping from "bad regions" suggests the vanishing gradient problem. There are two sources of gradient in the network: image loss ($L_P \& L_R$) and secret loss ($L_M$). The image loss is calculated after the encoder, and the secret loss is calculated after the decoder, so we log the gradients with respect to the first layer of the encoder and the first layer of the decoder, which are the farthest from two sources of gradient (Figure \ref{gradFigure}).We use the average of the absolute values of the gradients to represent the gradients of the whole layer. The experiments show that the gradient of StegaStamp is really small when it is caught in the "bad region". In particular, the gradient of the first layer of the decoder even drops to 0, which means that the vanishing gradient problem does occur.

In order to solve the vanishing gradient problem, we insert normalization layers before activation layers. As shown in the figures, the gradient of the network is effectively improved (Figure \ref{gradFigure}) and the model can stably and quickly reach close to 100\% decoding accuracy (Figure \ref{lossFigure}). We tried both batch normalization \cite{DBLP:journals/corr/IoffeS15} and instance normalization \cite{DBLP:journals/corr/UlyanovVL16}, both of which can improve the model, but instance normalization is less dependent on batch size and more effective, so we choose instance normalization. We also test where the normalization layer has the greatest impact on training.  The result is that just by removing the normalization layer before the final sigmoid activation layer of the decoder, the training is greatly affected. Although it is still possible to escape the "bad region", it is much slower. The result is also consistent with the fact that the gradient of the decoder is most affected by the vanishing gradient when no normalization layer is added.
\subsubsection{Self-Adaptive Weight}
In order to satisfy both image fidelity and information fidelity, the model needs to optimize both image loss and secret loss, so it is a multi-task learning. Typically for multi-task learning, tasks tend to be similar and can be mutually reinforced. However, image fidelity and information fidelity are inherently contradictory, which makes training difficult and requires a good balance of loses' weights. In addition, the increase of the strength of perturbations causes the input distribution of decoder to change continuously, and the training difficulty gradually increases, so the loses' weights also need to change accordingly.

Since information fidelity is harder to learn than image fidelity, it is necessary to give a larger weight to information fidelity in the early stage. To prioritize information fidelity, StegaStamp forces the image loss' weight as 0 before 1500 steps, then let the weights linearly grow between 1500 steps and 15000 steps, and finally keep them constant after 15000 steps. This introduces too many hyperparameters into training: When do weights start to grow? When do weights end increasing? To what extent do the weights increase? Adjusting parameters takes a lot of time and effort.
 
 \begin{figure}
    \centering
    \includegraphics[width=.5\textwidth]{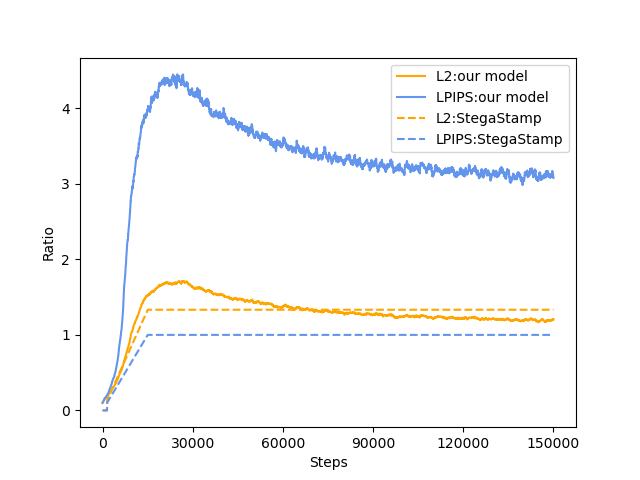}
    \caption{The change of losses' weights during training. Because the absolute value of weight is not important, we log the relative ratio between two kinds of image losses and secret loss. In StegaStamp, the ratio is 0 before 1500 steps, but it is too small to be noticeable on the figure.}
    \label{weightFigure}
\end{figure}
 
In order to reduce the number of hyperparameters, we use the method in the paper \cite{Kendall_2018_CVPR} to train the weights as learnable parameters, so that only the initial weights need to be set as hyperparameters. Just by setting the initial value of the secret loss' weight larger than the initial value of the image loss' weight, it can ensure the priority of training information fidelity. Self-adaptive weight not only reduces the number of hyperparameters, but also achieves better weight control. As shown in Figure \ref{weightFigure}, in StegaStamp, the image loss' weight to secret loss' weight ratio always increases , but with self-adaptive weight, the ratio first increases and then decreases after around 20k steps. That is the time when the strength of perturbations peaks, and the input distribution of the decoder is fixed. This automatic decrease of ratio helps the convergence of the model and the increase of decoding accuracy.

\section{Conclusion}
Our work explores and improves StegaStamp as a representation of deep watermark. By experiment, we reveal StegaStamp’s amazing robustness and its vulnerability to cropping. An interesting finding is that the vulnerability of cropping comes from the movement of the watermark’s position, but not the loss of information on the edge. We improve the robustness against cropping by introducing cropping into the training. Another shortcoming of StegaStamp is the difficulty of training. We introduce instance normalization to solve the vanishing gradient, set losses' weights as learnable parameters to reduce the number of hyperparameters, and use sigmoid to restrict pixel values of the generated image. The improvement of the resistance to cropping and the ease of training make deep watermark possible to be applied to a wider range of scenarios

\bibliographystyle{plain}
\bibliography{ref}

\end{document}